\documentclass[journal]{IEEEtran}
\usepackage{amsfonts}
\usepackage{amsmath}
\usepackage{amssymb}
\usepackage[numbers,sort&compress]{natbib}
\usepackage[colorlinks, linkcolor=red, anchorcolor=blue, citecolor=green]{hyperref}
\usepackage{bm}
\usepackage{graphicx}
\usepackage[dvipsnames]{xcolor}
\usepackage{booktabs}
\usepackage{subfigure}
\usepackage{multirow}

\ifCLASSINFOpdf
\else
\fi

\begin{document}

\title{Learning Spectral-Spatial-Temporal Features via a Recurrent Convolutional Neural Network for Change Detection in Multispectral Imagery}
\author{Lichao~Mou,~\IEEEmembership{Student Member,~IEEE,}
       ~Lorenzo~Bruzzone,~\IEEEmembership{Fellow,~IEEE,} and
       ~Xiao~Xiang~Zhu,~\IEEEmembership{Senior Member,~IEEE}

\thanks{Manuscript received March 6, 2018.}
\thanks{This work is jointly supported by the China Scholarship Council, the European Research Council (ERC) under the European Union¡¯s Horizon 2020 research and innovation programme (grant agreement No [ERC-2016-StG-714087], Acronym: \textit{So2Sat}), and Helmholtz Association under the framework of the Young Investigators Group ``SiPEO'' (VH-NG-1018, www.sipeo.bgu.tum.de).}
\thanks{L. Mou and X. X. Zhu are with the Remote Sensing Technology Institute (IMF), German Aerospace Center (DLR), Germany and with Signal Processing in Earth Observation (SiPEO), Technical University of Munich (TUM), Germany (e-mails: lichao.mou@dlr.de; xiao.zhu@dlr.de).}
\thanks{L. Bruzzone is with the Department of Information Engineering and Computer Science, University of Trento, Italy (e-mail: lorenzo.bruzzone@unitn.it).}
       }

\markboth{Submitted to IEEE Transactions on Geoscience and Remote Sensing on March 6, 2018}%
{Shell }

\maketitle

\begin{abstract}
Change detection is one of the central problems in earth observation and was extensively investigated over recent decades. In this paper, we propose a novel recurrent convolutional neural network (ReCNN) architecture, which is trained to learn a joint spectral-spatial-temporal feature representation in a unified framework for change detection in multispectral images. To this end, we bring together a convolutional neural network (CNN) and a recurrent neural network (RNN) into one end-to-end network. The former is able to generate rich spectral-spatial feature representations, while the latter effectively analyzes temporal dependency in bi-temporal images. In comparison with previous approaches to change detection, the proposed network architecture possesses three distinctive properties: 1) It is end-to-end trainable, in contrast to most existing methods whose components are separately trained or computed; 2) it naturally harnesses spatial information that has been proven to be beneficial to change detection task; 3) it is capable of adaptively learning the temporal dependency between multitemporal images, unlike most of algorithms that use fairly simple operation like image differencing or stacking. As far as we know, this is the first time that a recurrent convolutional network architecture has been proposed for multitemporal remote sensing image analysis. The proposed network is validated on real multispectral data sets. Both visual and quantitative analysis of experimental results demonstrates competitive performance in the proposed mode.
\end{abstract}


\IEEEpeerreviewmaketitle

\section{Introduction}
\label{sec:intro}
\IEEEPARstart{W}{ith} the development of remote sensing technology, every day massive amounts of remotely sensed data are produced from a rich number of spaceborne and airborne sensors; e.g., the Landsat 8 satellite is capable of imaging the entire Earth every 16 days in an eight-day offset from Landsat 7, and every 10 days the Sentinel-2 mission can provide a global coverage of the Earth's land surface. For the Sentinel-2 mission alone, to date about 3.4 petabytes of data have been acquired. Triggered by these exciting existing and future observation capabilities, methodological research on the multitemporal data analysis is of great importance~\cite{Bovolo15,Yokoya17}. Change detection is very crucial in the field of multitemporal image analysis, as it is able to identify land use or land cover differences in the same geographical area across a period of time and can be used in a large number of applications, to name a few, urban expansion, disaster assessment, resource management, and monitoring dynamics of land use~\cite{Yang12,Xian10,Liang11}.
\par
In the literature, many methods have been proposed to better identify land cover changes~\cite{Bovolo15}. Among them, a widely used model is based on image algebra approaches. A classic one is change vector analysis (CVA) proposed by Malila in 1980~\cite{Malila80}. CVA is designed to analyze possible multiple changes in pairs of multi-spectral pixels of bi-temporal images. Bovolo and Bruzzone~\cite{Bovolo07} propose a formal definition and a theoretical study to of CVA in the polar domain. Later some extensions of the CVA model have been proposed, e.g., compressed CVA (C$^2$VA)~\cite{Bovolo12}. CVA is used together with unsupervised threshold selection techniques based on different possible models of the data distribution. For example, the Rayleigh-Rice mixture density model~\cite{Zanetti15} has been recently used in the framework of the Expectation-Maximization (EM) algorithm.
\par
In addition, some image transformation-based models have been proposed in change detection to improve detection performance. These approaches mainly aim at learning a new, transformed feature representation from the original spectral domain, in order to suppress unchanged regions and highlight the presence of changes in the new feature space. For example, principal component analysis (PCA), Gram-Schmidt transformation, multivariate alteration detection (MAD), slow feature analysis (SFA), sparse learning, and deep belief network (DBN) use transformation algorithms in change detection methods~\cite{Deng08,Collins96,Nielsen98,Wu14,Erturk16,Gong17}. PCA is one of the best known subspace learning algorithms and can be used on both difference images and stacked images~\cite{Deng08,Li98}.
The goal of Gram-Schmidt transformation is to reduce data correlation.
MAD makes an attempt at maximizing variance of independently transformed variables~\cite{Nielsen98} and is invariant to linear scaling of the input data.
SFA~\cite{Wu14} is able to extract the most temporally invariant component from multitemporal images to transform data into a new feature space and, in this space, differences in unchanged pixels are suppressed so that changed regions can be better separated. In~\cite{Erturk16}, the authors apply sparse learning on stacked multitemporal images and expect that resulting sparse solutions do not vary greatly between the multitemporal data. In~\cite{Gong17}, the authors learn feature representations of two images with DBNs. Feature vectors issued from the two image acquisitions are stacked and used to learn a representation, where changes stand out more clearly. Using such feature representation, changes are more easily detected by image differencing.
\par
Another important branch of change detection methods is based on classification approaches. For example, Bruzzone and Serpico~\cite{Bruzzone97} propose a supervised nonparametric model, based on the compound classification rule for minimum error, to detect land cover transitions between two remote sensing images acquired at different times. The main idea of this approach is to consider the temporal correlation between images in the classification without requiring complex training data. In~\cite{Bruzzone99}, the authors use the Bayes rule for minimum error in the compound classification framework for detecting land cover transitions between pairs multisource images gathered at two different dates. In~\cite{Bruzzone02}, the authors propose a multi-classifier architecture, which is composed of an ensemble of partially unsupervised classifiers, to detect changes or update land cover maps. Later, Bruzzone et al.~\cite{Bruzzone04} develop an effective system that employs an ensemble of nonparametric multitemporal classifiers to address the problem of detecting land cover transitions in multitemporal images. All these techniques consider different tradeoffs between modeling the temporal correlation in the training of the system and requiring complex training data.
\par
One crucial issue in change detection is modeling the temporal correlation between bi-temporal images. Various atmospheric scattering conditions, complicated light scattering mechanisms, and intra-class variability lead to change detection being inherently nonlinear. Thus sophisticated, task-driven, learning-based methods are desirable.
\par
Deep neural networks have recently been shown to be very successful on a variety of computer vision and remote sensing tasks~\cite{Zhu17DLinRS,DLinRS}. They can also provide the opportunity for change detection, where one would like to extract joint spectral-temporal features from a bi-temporal image sequence in an end-to-end manner. In this respect, as an important branch of deep learning family, a recurrent neural network (RNN) is a natural candidate to tackle the temporal connection between multitemporal sequence data in change detection tasks. Recently, Lyu et al.~\cite{Lyu16} make use of an end-to-end RNN to solve the multi/hyper-spectral image change detection task, since RNN is well known to be good at processing sequential data. In their framework, a long short-term memory (LSTM)-based RNN is employed to learn a joint spectral-temporal feature representation from a bi-temporal image sequence. In addition, the authors also show the versatility of their network by applying it to detect multi-class changes and pointing out a good transferability for change detection in an ``unseen'' scene without fine-tuning. The authors of~\cite{Russwurm17} follow a similar idea, where an RNN based on LSTM units is used to extract dynamic spectral-temporal features but, in contrast to the change detection scenario, their goal is to address land cover classification of multitemporal image sequence.
\par
In this paper, we would like to learn joint spectral-spatial-temporal features using an end-to-end network for change detection, which is named as recurrent convolutional neural network (ReCNN), since it combines convolutional neural network (CNN) and RNN. Although both CNN~\cite{Mou18,Hujurse17,Moujurse17,Vol16,MouConvDeconvTGRS17,Maggiori17,Moudfc16,Hughes18,DFC16} and RNN~\cite{Lyu16,Mournn,Russwurm17} are well-established techniques for remote sensing applications, to the best of our knowledge, we are the first to combine them for multitemporal data analysis in the remote sensing community. Note that integrating CNN and RNN in an end-to-end manner has also been explored in hyperspectral image classification~\cite{Wu17}, where the network is only used for extracting spectral information to build a spectral classifier for the classification purpose. In our work, the CNN part transforms the input, a pair of 3D multispectral patches, to an abstract spectral-spatial feature representation, whereas the RNN part is not only employed for modeling temporal dependency, but is also used for predicting the final label (i.e., changed, unchanged, or change-type). In other words, the features from the proposed ReCNN encapsulate information related to spectral, spatial, and temporal components in bi-temporal images, making them useful for an holistic change detection task. For multitemporal image analysis, the proposed ReCNN contributes to the literature in three major aspects:
\begin{itemize}
  \item It is able to extract a spectral-spatial-temporal feature representation of multitemporal data through learning with a structured deep architecture.
  \item It has the same property of 2D CNN used for multi/hyper-spectral data classification on learning informative spectral-spatial feature representations directly from multispectral data, requiring neither hand-crafted visual features nor pre-processing steps.
  \item It has the same characteristic of RNN, being capable of modeling the temporal correlation between bi-temporal images using a sophisticated and task-driven approach to the extraction of temporal features in an end-to-end architecture, and finally producing labels for the image sequence.
\end{itemize}
\par
The remainder of this paper is organized as follows. After the introductory Section~\ref{sec:intro} detailing change detection, Section~\ref{sec:med} is dedicated to the details of the proposed recurrent convolutional network. Section~\ref{sec:exp} then provides data set information, network setup, experimental results, and discussion. Finally, Section~\ref{sec:con} concludes the paper.

\begin{figure*}[!t]
\centering
\includegraphics[width=\linewidth]{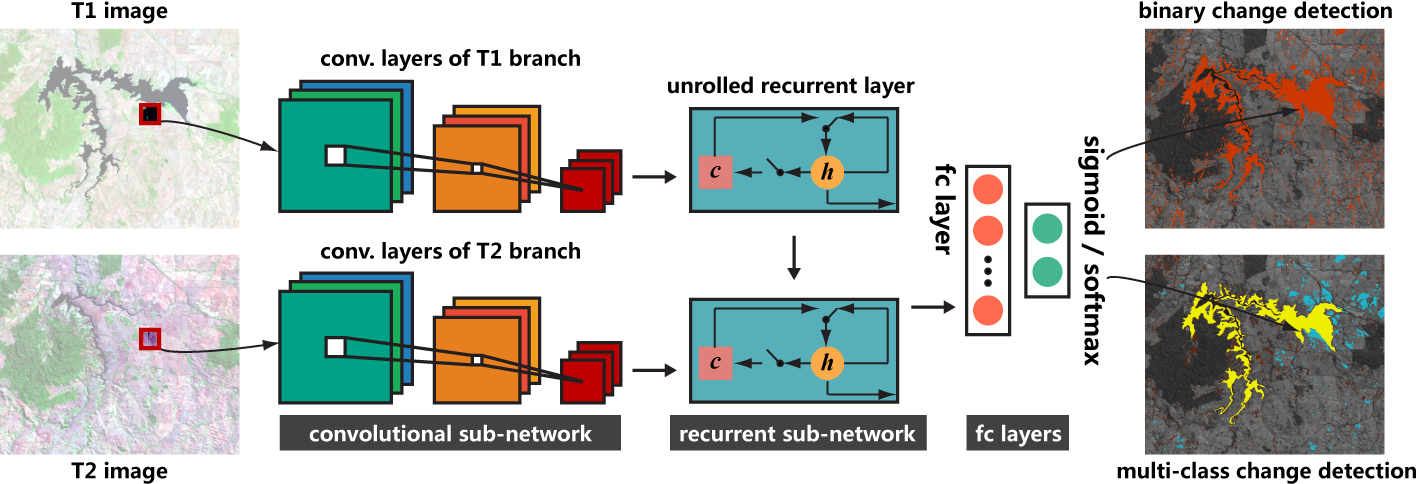}
\renewcommand{\figurename}{Fig}
\caption{\label{fig:net} Overview of the proposed recurrent convolutional neural network (ReCNN). At the bottom of our network, convolutional layers automatically extract feature maps from each input. On top of the convolutional sub-network, a recurrent sub-network takes the feature representations produced by convolutional layers as inputs to exploit the temporal dependency in the bi-temporal images. The third part is two fully connected layers widely used in classification problems. Although ReCNN is composed of different kinds of network architectures, i.e., CNN, RNN, and fully connected network, it can be trained end-to-end by back-propagation with one loss function, due to the differential property of all these components.}
\end{figure*}

\section{Methodology}
\label{sec:med}
\subsection{Network Architecture}
The architecture of the proposed ReCNN, as shown in Fig.~\ref{fig:net}, is made up of three components, including a convolutional sub-network, a recurrent sub-network, and fully connected layers, from bottom to top.
\par
To acquire a joint spectral-spatial-temporal feature representation for change detection, at the bottom of our network, convolutional layers automatically extract feature maps from each input. On top of the convolutional sub-network, a recurrent sub-network takes the feature representations produced by convolutional layers as inputs to exploit the temporal dependency in the bi-temporal images. The third part is two fully connected layers widely used in classification problems. Although ReCNN is composed of different kinds of network architectures (i.e., CNN, RNN, and fully connected network) it can be trained end-to-end by back-propagation with one loss function, due to the differential properties of all these components.
\par
Let $\bm{X}^{T_1}$ and $\bm{X}^{T_2}$ represent a pair of multispectral images acquired over the same geographical area at two different times $T_1$ and $T_2$, respectively. Let $\bm{x}^{T_1}$ and $\bm{x}^{T_2}$ be two patches taken from the exact same location in two images. $\bm{y}$ is a label that indicates the category (i.e., changed, unchanged, or change-type) that the pair of patches belongs to. The flowchart of the proposed ReCNN can be summarized as follows:
\begin{itemize}
  \item First, the 3D multispectral patch $\bm{x}^{T_1}$ is fed into $T_1$ branch of the convolutional sub-network, which transforms it to an abstract feature vector $\bm{f}^{T_1}$.
  \item Then, the recurrent sub-network receives $\bm{f}^{T_1}$ and calculates the hidden state information for the current input; it also restores that information in the meantime.
  \item Subsequently, $\bm{x}^{T_2}$ is input to $T_2$ branch for extracting spectral-spatial feature $\bm{f}^{T_2}$, it is fed into the recurrent layer simultaneously with the state information of $\bm{f}^{T_1}$, and the activation at time $T_2$ is computed by a linear interpolation between existing value and the activation of the previous time $T_1$.
  \item Finally, fully connected layers of the ReCNN predict the label of the input bi-temporal multispectral patches by looping through the entire sequence.
\end{itemize}
\par
The entire change detection map can be obtained by applying the network to all pixels in the image.

\begin{figure*}[!t]
\centering
\includegraphics[width=0.7\linewidth]{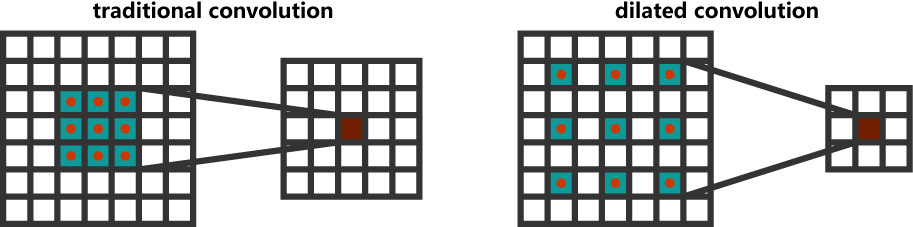}
\renewcommand{\figurename}{Fig}
\caption{\label{fig:conv} Illustration of traditional convolution operation (left) and 2-dilated convolution (right). Traditional convolution corresponds to dilated convolution with dilation rate 1. Employing dilated convolution operation enlarges the network's field of view.}
\end{figure*}

\subsection{Spectral-Spatial Feature Extraction via the Convolutional Sub-Network}
As we have mentioned, spectral-spatial information is of great importance for change detection. Some of the previous widely used unsupervised image algebra-based and image transformation-based methods cannot totally capture task specialized features which may be discriminative for a specific change detection task. Features directly learned from data and driven by tasks are supposed to be better~\cite{DLinRS}. This advantage leads to our usage of a trainable feature generator.
\par
Though trainable, early and fairly simple 1D neural network models, such as DBN~\cite{Gong17} and multilayer perceptron (MLP), suffer from huge amount of learnable parameters, since those architectures are totally equipped with fully connected layers, which is an undesirable case given that available annotated training samples for change detection are often very limited. Moreover, another disadvantage of such networks is that they treat the multispectral data as vectors, ignoring the 2D property of imagery in the spatial domain.
\par
CNNs, which are a significant branch of deep learning, have been attracting attention, due to the fact that they are capable of automatically discovering relevant contextual 2D spatial features as well as spectral features for multi/hyper-spectral data. In addition, a CNN makes use of local connections to deal with spatial dependencies via sharing weights, and thus can significantly reduce the number of parameters of the network in comparison with the conventional 1D fully connected neural networks, e.g., DBN and MLP. Recently, CNNs used for hyperspectral image classification have proven their effectiveness in extracting useful spectral-spatial features~\cite{ShenRS16,MouConvDeconvTGRS17}. Triggered by this, adopting a CNN in our architecture is natural.
\par
However, a direct use of CNNs commonly used in typical recognition tasks, e.g., AlexNet~\cite{AlexNet}, VGG Nets~\cite{VGGNet}, and GoogLeNet~\cite{GoogLeNet}, is not possible in our task, as we believe that a simpler network architecture is more appropriate for our problem due to the following reasons. First, change detection aims to distinguish only several classes (two for binary change detection), which requires much less model complexity than general visual recognition problems in computer vision, such as ImageNet classification with 1,000 categories. Second, since spatial resolution of multispectral imagery is limited, it is desirable to make input size small, which reduces the depth of the network naturally. Third, a smaller network is obviously more efficient in change detection problems, where testing may be performed in a large-scale area. Finally, the above-mentioned networks are not suitable to be used on multispectral images with a large number of spectral channels.
\par
The convolutional sub-network receives a sequence of $5\times 5$ multispectral patches as the input and has two separate, yet identical convolutional branches (i.e., $T_1$ branch and $T_2$ branch [cf. Fig.~\ref{fig:net}]) which process $\bm{x}^{T_1}$ and $\bm{x}^{T_2}$ in parallel, respectively. The learned features are fed into the following recurrent sub-network. Using this two-branch architecture, the convolutional recurrent neural network is constrained to first learn meaningful spectral-spatial representations of input patches, and to combine them on a higher level for modeling temporal dependency. More specifically, we make use of convolutional filters with a very small receptive field of $3\times 3$, rather than using larger ones such as $5\times 5$. Moreover, we do not adopt max-pooling after convolution or spatial padding for convolutional layers. The depth of the convolutional sub-network is such that the output size of the last layer is $1\times 1$.
\par
Regarding convolution, we make use of dilated convolution to construct convolutional layers in the network because, for our task, it is able to offer a slightly better performance than a traditional convolution operation. The dilated convolution~\cite{deeplab} was originally designed for the efficient computation of the undecimated wavelet transform in the ``algorithme \`a trous'' scheme~\cite{Holschneider89}. This algorithm makes it possible to calculate responses of any layer at any desirable resolution and can be applied post-hoc, once a network has been trained. Let $F:\mathbb{Z}^2\rightarrow \mathbb{R}$ be a discrete function. Let $\Omega_r=[-r,r]^2\cap\mathbb{Z}^2$ and let $k:\Omega_r\rightarrow\mathbb{R}$ be a discrete filter of size $(2r+1)^2$. The traditional discrete convolution operation $\ast$ can be defined as follows:
\begin{equation}
(F\ast k)({\rm p})=\sum_{{\rm s+t=p}}F({\rm s})k({\rm t})\,.
\end{equation}
\par
This operation can be generalized. Let $l$ be a dilation rate and let $\ast_l$ be defined as
\begin{equation}
(F\ast_l k)({\rm p})=\sum_{{\rm s}+l{\rm t=p}}F({\rm s})k({\rm t})\,.
\end{equation}
\par
We will refer to $\ast_l$ as a dilated convolution or an $l$-dilated convolution. Fig.~\ref{fig:conv} shows differences between the conventional convolution and the dilated convolution.
\par
The usage of dilated convolution in our network allows us to exponentially enlarge the field of view with linearly increasing number of parameters, providing a significant parameter reduction while increasing effective field of view. Note that a very recent study~\cite{Peng17} found that large field of view actually plays an important role. This can be easily understood by an analogy that states the fact that humans usually confirm the category of a pixel by referring to its surrounding context region.

\begin{figure*}[!t]
\centering
\includegraphics[width=\linewidth]{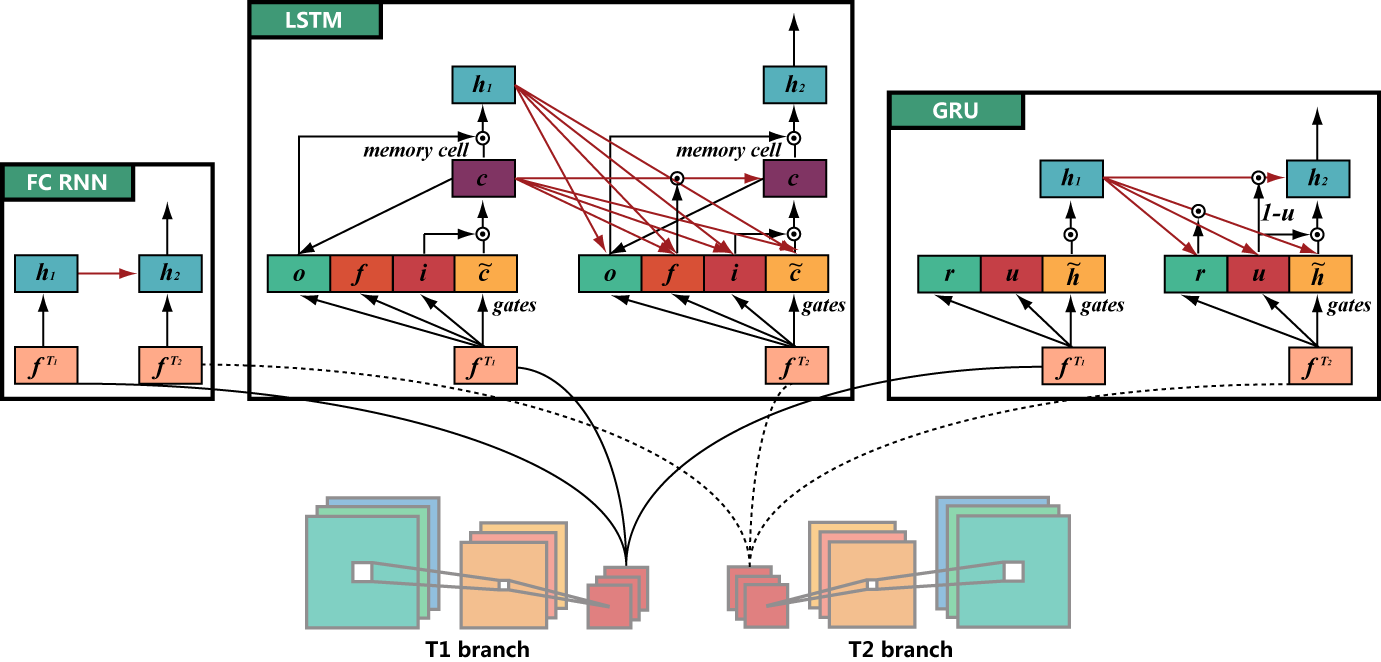}
\renewcommand{\figurename}{Fig}
\caption{\label{fig:rnn} Graphic models of fully connected RNN, LSTM, and GRU. In LSTM, $\bm{o}$, $\bm{f}$, $\bm{i}$, $\tilde{\bm{c}}$, and $\bm{c}$ are output gates, forget gates, input gates, new memory cell contents, and memory cells, respectively. In GRU, the reset and update gates are denoted by $\bm{r}$ and $\bm{u}$, and $\tilde{\bm{h}}$ and $\bm{h}$ are the candidate activation and final activation.}
\end{figure*}

\subsection{Modeling Temporal Dependency by the Recurrent Sub-Network}
The impressive success of recent deep learning systems has been predominantly achieved by feedforward neural network architectures like CNN. In such networks, we implicitly assume that all inputs are independent of each other. However, for tasks that involve processing time sequence (e.g., change detection), that is not a good assumption. RNNs are a kind of neural networks that extend the conventional feedforward neural networks with loops in connections. Unlike a feedforward network, an RNN is capable of dealing with dependent, sequential inputs by having a recurrent hidden state whose activation at each time step depends on that of the previous time. By doing so, the network can exhibit dynamic temporal behavior, which is in line with our purpose; i.e., modeling temporal dependency between the $T_1$ and $T_2$ data. To this end, three types of RNN architectures, namely, fully connected RNN, LSTM, and gated recurrent unit (GRU), are used to construct the recurrent sub-network in our ReCNN.
\par
\textbf{Fully Connected RNN.} Given feature vectors $\bm{f}^{T_1}$ and $\bm{f}^{T_2}$ learned from the convolutional sub-network, a fully connected RNN updates its recurrent hidden state $\bm{h}_t$ by
\begin{equation}\label{eq:rnn1}
\bm{h}_t=
\begin{cases}
0 &\mbox{if $t=0$}\\
\varphi(\bm{h}_{t-1},\bm{f}^{T_t}) &\mbox{otherwise}
\end{cases}\,,
\end{equation}
where $\varphi$ is a nonlinear activation function, such as a hyperbolic tangent function or logistic sigmoid function. The recurrent layer will output a sequence $\bm{h}=(\bm{h}_1, \bm{h}_2)$. For our task, we only need the last one as input to the fully connected layers for predicting label.
\par
In the fully connected RNN model, the update of the recurrent hidden state in Eq.~(\ref{eq:rnn1}) is implemented as
\begin{equation}\label{eq:rnn2}
\bm{h}_t=\varphi(\bm{U}\bm{h}_{t-1}+\bm{W}\bm{f}^{T_t})\,,
\end{equation}
where $\bm{U}$ and $\bm{W}$ are the coefficient matrices for the activation of recurrent hidden units at the previous time step and for the input at the present time, respectively.
\par
Fully connected RNN is the concisest RNN model, and it can reflect the essence of RNNs; i.e., an RNN is capable of modeling a probability distribution over the next element of the sequence data, given its present state $\bm{h}_t$, by capturing a distribution over sequence data. Let $p(\bm{f}^{T_1},\bm{f}^{T_2})$ be the sequence probability, which can be decomposed into
\begin{equation}
p(\bm{f}^{T_1},\bm{f}^{T_2})=p(\bm{f}^{T_1})p(\bm{f}^{T_2}|\bm{f}^{T_1})\,.
\end{equation}
\par
Then, the conditional probability distribution can be modeled with an RNN:
\begin{equation}
p(\bm{f}^{T_2}|\bm{f}^{T_1})=\varphi(\bm{h}_2)\,,
\end{equation}
where $\bm{h}_2$ is obtained from Eq.~(\ref{eq:rnn1}). Our motivation in this work is apparent here: bi-temporal images act as true sequential data instead of a simple difference image or stacked image and, therefore, an RNN can be used to model the temporal dependency.
\par
\textbf{LSTM.} LSTM is a special type of recurrent hidden unit and was initially proposed by Hochreiter and Schmidhuber~\cite{lstm1}. Since then, a couple of minor modifications to the original version have been made. In this work, we follow the implementation of LSTM as used in~\cite{lstm2}. As shown in Eq.~(\ref{eq:rnn1}), recurrent hidden units in a fully connected RNN simply compute a weight sum of inputs and then apply a nonlinear function. In contrast, an LSTM-based recurrent layer maintains a series of memory cells $\bm{c}_t$ at time step $t$. The activation of LSTM units can be calculated by
\begin{equation}
\bm{h}_t=\bm{o}_t\tanh(\bm{c}_t)\,,
\end{equation}
where $\tanh(\cdot)$ is the hyperbolic tangent function and $\bm{o}_t$ is the output gates that control the amount of memory content exposure. The output gates are updated by
\begin{equation}
\bm{o}_t=\sigma(\bm{W}_{oi}\bm{f}^{T_t}+\bm{W}_{oh}\bm{h}_{t-1}+\bm{W}_{oc}\bm{c}_t)\,,
\end{equation}
where the $\bm{W}$ terms represent coefficient matrices; e.g., $\bm{W}_{oi}$ and $\bm{W}_{oc}$ are the input-output weight matrix and memory-output weight matrix, respectively.
\par
The memory cells $\bm{c}_t$ are updated by partially discarding the present memory contents and adding new contents of the memory cells $\tilde{\bm{c}}_t$:
\begin{equation}\label{eq:lstm}
\bm{c}_t=\bm{i}_t\odot\tilde{\bm{c}}_t+\bm{f}_t\odot\bm{c}_{t-1}\,,
\end{equation}
where $\odot$ is an element-wise multiplication. The new memory contents are
\begin{equation}
\tilde{\bm{c}}_t=\tanh(\bm{W}_{ci}\bm{f}^{T_t}+\bm{W}_{ch}\bm{h}_{t-1})\,,
\end{equation}
where $\bm{W}_{ci}$ is input-memory weight matrix and $\bm{W}_{ch}$ represents hidden-memory coefficient matrix.
\par
The $\bm{i}_t$ and $\bm{f}_t$ are input gates and forget gates, respectively. The former modulates the extent to which the new memory information is added to the memory cell, whereas the latter controls the degree to which contents of the existing memory cells are forgotten. Specifically, gates are computed as follows:
\begin{equation}
\bm{i}_t=\sigma(\bm{W}_{ii}\bm{f}^{T_t}+\bm{W}_{ih}\bm{h}_{t-1}+\bm{W}_{ic}\bm{c}_{t-1})\,,
\end{equation}
\begin{equation}
\bm{f}_t=\sigma(\bm{W}_{fi}\bm{f}^{T_t}+\bm{W}_{fh}\bm{h}_{t-1}+\bm{W}_{fc}\bm{c}_{t-1})\,.
\end{equation}

\begin{figure*}[!t]
\centering
\includegraphics[width=\linewidth]{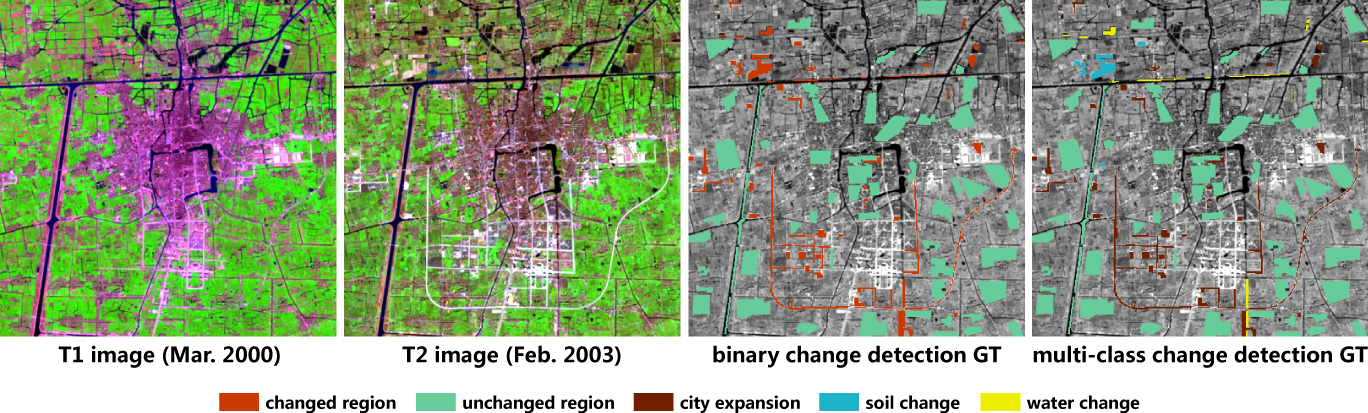}
\renewcommand{\figurename}{Fig}
\caption{\label{fig:taizhou} True-color composites of the $T_1$ and $T_2$ images in the Taizhou data set as well as ground truths (GTs).}
\end{figure*}
\par
\textbf{GRU.} Similarly to LSTM, a GRU makes use of a linear sum between the existing state and the newly computed state. It, however, directly exposes whole state values at each time step, instead of controlling what part of the state information will be exposed.
\par
The activation $\bm{h}_t$ of GRUs at time step $t$ is a linear interpolation between the previous activation $\bm{h}_{t-1}$ and the candidate activation $\tilde{\bm{h}}_t$:
\begin{equation}\label{eq:rgu}
\bm{h}_t=(\bm{1}-\bm{u}_t)\bm{h}_{t-1}+\bm{u}_t\tilde{\bm{h}}_t\,,
\end{equation}
where the update gates $\bm{u}_t$ determine how much GRUs update their activations or contents. Update gates can be computed by
\begin{equation}
\bm{u}_t=\sigma(\bm{W}_{ui}\bm{f}^{T_t}+\bm{W}_{uh}\bm{h}_{t-1})\,,
\end{equation}
where $\bm{W}_{ui}$ and $\bm{W}_{uh}$ are the input-update coefficient matrix and hidden-update weight matrix, respectively.
\par
The candidate activation $\tilde{\bm{h}}_t$ is computed similarly to that of the fully connected RNN (cf. Eq.~(\ref{eq:rnn1})) and as follows
\begin{equation}
\tilde{\bm{h}}_t=\tanh(\bm{U}(\bm{r}_t\odot\bm{h}_{t-1})+\bm{W}\bm{f}^{T_t})\,,
\end{equation}
where $\bm{r}_t$ is the set of reset gates. When reset gates are totally off (i.e., $\bm{r}_t$ is $\bm{0}$), GRUs will completely forget the activation of the recurrent layer at previous time and only receive existing input. When open, reset gates will partially keep the information of the previously computed state. Reset gates are calculated similarly to update gates:
\begin{equation}
\bm{r}_t=\sigma(\bm{W}_{ri}\bm{f}^{T_t}+\bm{W}_{rh}\bm{h}_{t-1})\,,
\end{equation}
where $\bm{W}_{ri}$ is the input-reset weight matrix and $\bm{W}_{rh}$ represents the hidden-reset coefficient matrix.
\par
Fig.~\ref{fig:rnn} shows graphic models of fully connected RNN, LSTM, and GRU through time.

\subsection{Network Training}
The network training is based on the TensorFlow framework. We chose Nesterov Adam~\cite{nadam2,nadam1} as the optimizer to train the network since, for this task, it shows much faster convergence than standard stochastic gradient descent (SGD) with momentum~\cite{sgd} or Adam~\cite{adam}. We fixed almost all of parameters of Nesterov Aadam as recommended in~\cite{nadam2}: $\beta_1=0.9$, $\beta_2=0.999$, $\epsilon=1\mathrm{e}{-08}$, and a schedule decay of 0.004, making use of a fairly small learning rate of $2\mathrm{e}{-04}$. All network weights are initialized with a Glorot uniform initializer~\cite{Glorot_normal} that draws samples from a uniform distribution. We utilize sigmoid and softmax as activation functions of the last fully connected layer for the binary and multi-class change detections, respectively. Finally, we train our network on a single NVIDIA GeForce GTX TITAN with 12 GB of GPU memory.

\section{Experimental Results and Discussion}
\label{sec:exp}
\subsection{Data Description}
The performance of the proposed network is evaluated on two data sets, which were acquired by the Landsat Enhanced Thematic Mapper Plus (ETM+) sensor with six bands and a spatial resolution of 30m. Before feeding data into models, digital numbers (DNs) of the original data were converted into absolute radiance (i.e., all of the data sets used in the experiments were normalized into a range of [0,1]).
\par
\subsubsection{Taizhou Data}
This data set consists of two images covering the city of Taizhou, China, in March 2000 and February 2003, with a WGS-84 projection and a coordinate range of 31$^{\circ}$14$'$56N--31$^{\circ}$27$'$39N, 120$^{\circ}$02$'$24E--121$^{\circ}$07$'$45E. These two images both consist of $400\times400$ pixels, and changes between them mainly involve city expansion. The available manually annotated samples of this data set for multi-class change detection cover four classes of interest (cf. Fig.~\ref{fig:taizhou}); i.e., unchanged area, city expansion (bare soils, grasslands, or cultivated fields to buildings or roads), soil change (cultivated field to bare soil), and water change (non-water regions to water regions). Table~\ref{tab:taizhou} provides information about different classes and their corresponding training and test samples.

\begin{table}[h]
\caption{\label{tab:taizhou} Number of Training and Test Samples in the Taizhou Data Set}
\centering
\begin{tabular}{cccc}
\toprule[1pt]
 & \textbf{Class Name} & \textbf{Training} & \textbf{Test} \\
\toprule[0.5pt]
\multirow{3}{*}{\textbf{Binary}} & Changed region & 500 & 4055 \\
 & Unchanged region & 500 & 16961 \\
 & TOTAL & 1000 & 21016 \\
\toprule[0.5pt]
\multirow{5}{*}{\textbf{Multiple}} & Unchanged region & 500 & 16961 \\
 & City expansion & 500 & 2875 \\
 & Soil change & 500 & 104 \\
 & Water change & 500 & 75 \\
 & TOTAL & 2000 & 20015 \\
\bottomrule[1pt]
\end{tabular}
\end{table}


\subsubsection{Eppalock Lake}
\begin{figure*}[!t]
\centering
\includegraphics[width=\linewidth]{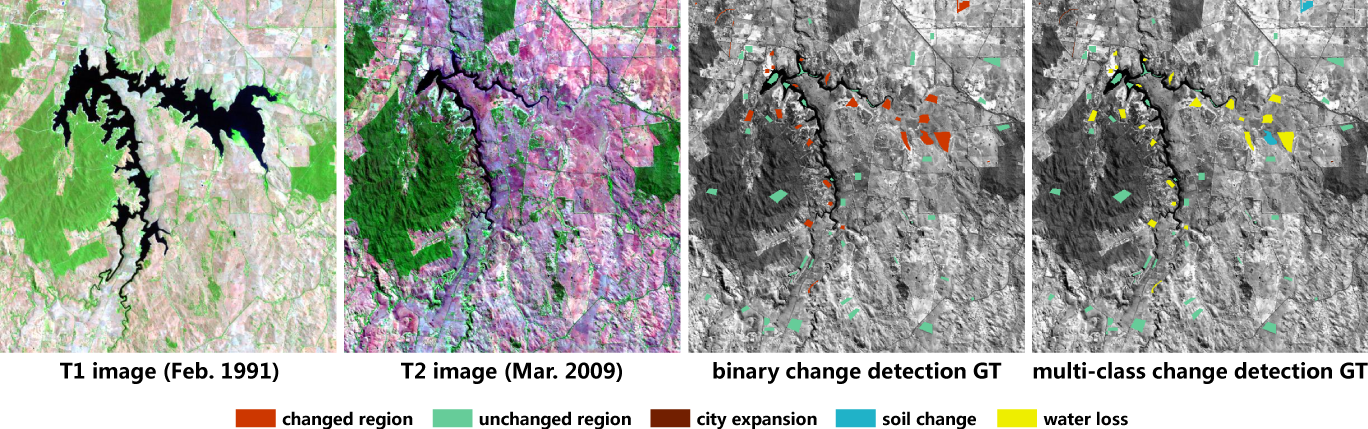}
\renewcommand{\figurename}{Fig}
\caption{\label{fig:eppalock} Eppalock lake data set.}
\end{figure*}

The second data set was acquired over the Eppalock lake, Victoria, Australia, in February 1991 and March 2009, with a WGS-84 projection and a coordinate range of 36$^{\circ}$49$'$10S--37$^{\circ}$00$'$52S, 144$^{\circ}$27$'$52E--144$^{\circ}$37$'$35E. Both images in this data set are $602\times631$ pixels. Similar to the Taizhou data, four multi-class change types are considered in the Eppalock lake scene, and they are unchanged region, city expansion (bare soils, grasslands, or cultivated fields to buildings or roads), water loss (water regions to bare soils), and soil change (vegetative covers or artificial buildings to bare soils). Fig.~\ref{fig:eppalock} shows tow true-color composite images and their corresponding reference samples. The number of training and test samples is displayed in Table~\ref{tab:lake}.

\subsection{General Information}
To evaluate the performance of different change detection algorithms, we utilize the following evaluation criteria:
\begin{itemize}
  \item Overall accuracy (OA): This index shows the number of bi-temporal pixels that are classified correctly, divided by the number of test samples.
  \item Kappa coefficient: This metric is a statistical measurement of agreement between the final change detection map and the ground truth map. It is the percentage agreement corrected by the level of agreement that could be expected due to change alone. In general, it is thought to be a more robust measure than a simple percent agreement computation, as $k$ takes into account the agreement occurring by chance.
\end{itemize}
\par
To validate the effectiveness of the proposed ReCNN model, it is compared with the most widely used change detection methods. These methods are summarized as follows:
\begin{itemize}
  \item CVA~\cite{Bovolo07}, which is an effective unsupervised approach for multispectral image change detection tasks.
  \item PCA~\cite{Deng08}, which is simple in computation and can be applied to real-time applications.
  \item MAD~\cite{Nielsen98}, which is a classical image transformation-based unsupervised algorithm for bi-temporal multispectral image change detection.
  \item Iteratively-reweighted multivariate alteration detection (IRMAD)~\cite{Nielsen07}, which is an extension to MAD by introducing an iterative scheme.
  \item Decision tree (DT), which is a non-parametric supervised learning method used for classification and regression. Its goal is to create a model that predicts the value of a target variable by learning simple decision rules inferred from data features.
  \item Support vector machine (SVM), which works by mapping data to a kernel-included high-dimensional feature space seeking an optimal decision hyperplane that can best separate data samples, when data points are not linearly separable. Here, we use an SVM with RBF kernel. The optimal hyperplane parameters $C$ (parameter that controls the amount of penalty during the SVM optimization) and $\gamma$ (spread of the RBF kernel) have been traced in the range of $C=10^{-2},10^{-1},\cdots,10^{4}$ and $\gamma=2^{-3},2^{-2},\cdots,2^{4}$ using five-fold cross validation.
  \item RNN~\cite{Lyu16}, which has recently shown promising performance in classification and change detection.
  \item ReCNN-FC, which uses fully connected RNN as recurrent sub-network in ReCNN model.
  \item ReCNN-GRU, which uses GRU architecture in the recurrent sub-network.
  \item ReCNN-LSTM, which is the ReCNN model with LSTM as recurrent component.
\end{itemize}
\par
Among these methods, CVA, PCA, MAD, IRMAD, and RNN are used in binary change detection experiments, and DT, SVM, and RNN are compared to the proposed network in multi-class change detection experiments. Moreover, k-means algorithm is used to automatically select threshold for unsupervised methods in the binary change detection task.

\begin{table}[t]
\caption{\label{tab:lake} Number of Training and Test Samples in the Eppalock Lake Data Set}
\centering
\begin{tabular}{cccc}
\toprule[1pt]
 & \textbf{Class Name} & \textbf{Training} & \textbf{Test} \\
\toprule[0.5pt]
\multirow{3}{*}{\textbf{Binary}} & Changed region & 500 & 3380 \\
 & Unchanged region & 500 & 4515 \\
 & TOTAL & 1000 & 7895 \\
\toprule[0.5pt]
\multirow{5}{*}{\textbf{Multiple}} & Unchanged region & 300 & 4715 \\
 & Water loss & 300 & 2817 \\
 & Soil change & 300 & 341 \\
 & City expansion & 50 & 72 \\
 & TOTAL & 950 & 7945 \\
\bottomrule[1pt]
\end{tabular}
\end{table}

\begin{figure}[t]
\centering
\includegraphics[width=0.7\columnwidth]{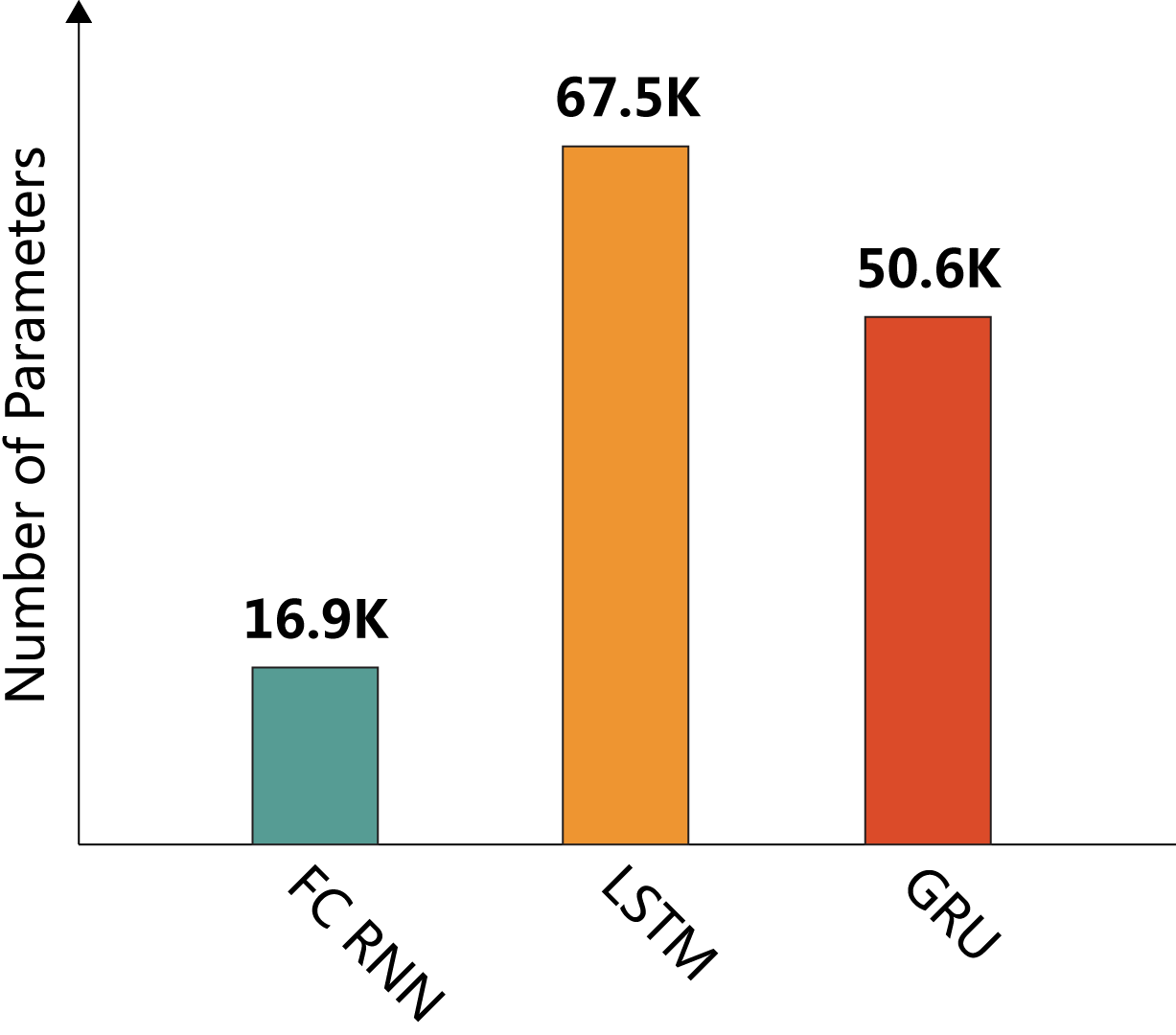}
\renewcommand{\figurename}{Fig}
\caption{\label{fig:rnn_sizes} Comparisons of different RNN architectures in terms of model size. Here, 128 recurrent units are used in each architecture.}
\end{figure}

\begin{table*}[!t]
\caption{\label{tab:binary} Accuracy Comparison of Binary Change Detection on the Two Experimental Data Sets.}
\centering
\begin{tabular}{cccccccccccc}
\toprule[1pt]
 & \multicolumn{4}{c}{\textbf{Taizhou City}} & & \multicolumn{4}{c}{\textbf{Eppalock Lake}} \\
 & OA & Kappa & Unchanged & Changed & & OA & Kappa & Unchanged & Changed \\
\toprule[0.5pt]
CVA~\cite{Bovolo07} & 83.82 & 0.3202 & 97.38 & 27.10 & & 81.28 & 0.6353 & 69.24 & 97.37 \\
PCA~\cite{Deng08} & 94.63 & 0.8181 & \textbf{99.79} & 74.51 & & 74.68 & 0.5044 & 64.98 & 87.63 \\
MAD~\cite{Nielsen98} & 94.62 & 0.8168 & 98.47 & 78.52 & & 91.10 & 0.8138 & 99.14 & 80.36 \\
IRMAD~\cite{Nielsen07} & 95.14 & 0.8313 & 99.35 & 77.53 & & 91.27 & 0.8174 & \textbf{99.49} & 80.30 \\
RNN~\cite{Lyu16} & 96.50 & 0.8884 & 97.58 & 91.96 & & 95.21 & 0.9018 & 97.03 & 92.78 \\
ReCNN-FC & 98.35 & 0.9470 & 98.94 & 95.86 & & 98.40 & 0.9674 & 98.56 & 98.20 \\
ReCNN-GRU & 98.67 & 0.9571 & 99.23 & 96.30 & & 98.64 & 0.9723 & 99.22 & 97.87 \\
ReCNN-LSTM & \textbf{98.73} & \textbf{0.9592} & 99.20 & \textbf{96.77} & & \textbf{98.67} & \textbf{0.9728} & 98.83 & \textbf{98.46} \\
\bottomrule[1pt]
\end{tabular}
\end{table*}

\begin{table*}[!t]
\caption{\label{tab:multi} Accuracy Comparison of Multi-Class Change Detection on the Two Experimental Data Sets.}
\centering
\begin{tabular}{cccccccc}
\toprule[1pt]
 &  & OA & Kappa & Unchanged & City expansion & Soil change & Water change/loss \\
\toprule[0.5pt]
\multirow{6}{*}{\textbf{Taizhou City}} & Decision Tree & 85.19 & 0.5846 & 84.64 & 88.49 & 82.69 & 86.67 \\
 & SVM & 93.90 & 0.7927 & 94.69 & 89.32 & 92.31 & 93.33 \\
 & RNN~\cite{Lyu16} & 95.48 & 0.8374 & 97.04 & 86.92 & 85.58 & 85.33 \\
 & ReCNN-FC & 97.37 & 0.9039 & 97.95 & 94.12 & \textbf{95.19} & 92.00 \\
 & ReCNN-GRU & 97.52 & 0.9097 & 98.05 & 94.54 & \textbf{95.19} & 96.00 \\
 & ReCNN-LSTM & \textbf{98.04} & \textbf{0.9279} & \textbf{98.36} & \textbf{96.31} & 94.23 & \textbf{97.33} \\
\toprule[0.5pt]
\multirow{6}{*}{\textbf{Eppalock Lake}} & Decision Tree & 87.56 & 0.7811 & 81.31 & 41.67 & 89.15 & 99.01 \\
 & SVM & 95.86 & 0.9228 & 94.46 & 72.22 & 97.65 & 98.58 \\
 & RNN~\cite{Lyu16} & 96.34 & 0.9392 & 95.55 & 41.67 & 96.48 & 99.04 \\
 & ReCNN-FC & 98.45 & 0.9705 & 98.01 & 80.56 & \textbf{100} & \textbf{99.47} \\
 & ReCNN-GRU & 98.49 & 0.9712 & 98.24 & 79.17 & \textbf{100} & 99.22 \\
 & ReCNN-LSTM & \textbf{98.70} & \textbf{0.9752} & \textbf{98.49} & \textbf{84.72} & \textbf{100} & 99.25 \\
\bottomrule[1pt]
\end{tabular}
\end{table*}

\subsection{Analysis of Recurrent Sub-network: Comparisons between Fully Connected RNN, LSTM, and GRU}
The most prominent trait shared between fully connected RNN, LSTM, and GRU is that there exists an additive loop of their update from $T_1$ to $T_2$, which is lacking in the conventional feedforward neural networks such as CNNs. In contrast, compared to the fully connected RNN like Eq.~(\ref{eq:rnn2}), both LSTM and GRU keep the current content and add the new content on top of it (cf. Eq.~(\ref{eq:lstm}) and Eq.~(\ref{eq:rgu})). These two RNN architectures, however, have a number of differences as well. LSTM makes use of three gates and a cell, namely, an input gate, forget gate, output gate, and memory cell, to control the exposure of memory content; whereas GRU only utilizes two gates to control the information flow. Therefore, the total number of parameters in GRU is reduced by about 25\% compared to that in LSTM. Fig.~\ref{fig:rnn_sizes} shows the number of total trainable parameters in different RNN architectures.
\par
Table~\ref{tab:binary} and Table~\ref{tab:multi} list binary and multi-class change detection results obtained in our experiments, respectively. For both data sets, ReCNN-LSTM outperforms ReCNN-FC and ReCNN-GRU on all indexes (i.e., OA and Kappa coefficient). For example, in the binary change detection, ReCNN-LSTM increases the accuracy by 0.38\% of OA and 0.0122 of Kappa on the Taizhou data set, in comparison with ReCNN-FC; by 0.06\% of OA and 0.0021 of Kappa on the same data set, compared to ReCNN-GRU. However, we can see that on these data sets, all three variations of the proposed ReCNN perform closely to each other. On the other hand, the proposed networks with gating RNN architectures as the recurrent sub-network (ReCNN-LSTM and ReCNN-GRU) slightly outperforms the more traditional ReCNN-FC on both of data sets and change detection tasks.

\begin{figure*}[!t]
\centering
\includegraphics[width=0.85\linewidth]{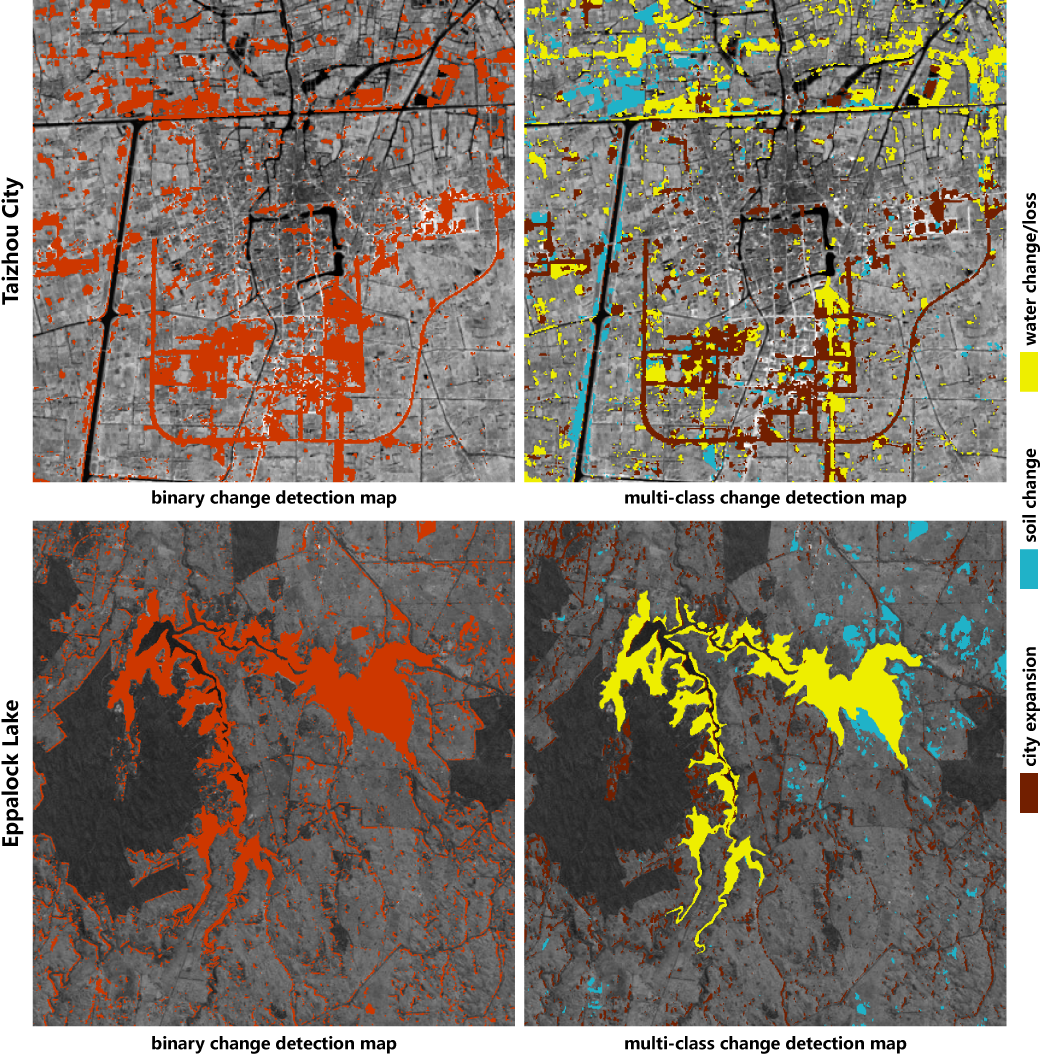}
\renewcommand{\figurename}{Fig}
\caption{\label{fig:change_maps} Change detection maps generated by the proposed ReCNN-LSTM model.}
\end{figure*}

\begin{figure}[!t]
\centering
\includegraphics[width=0.75\columnwidth]{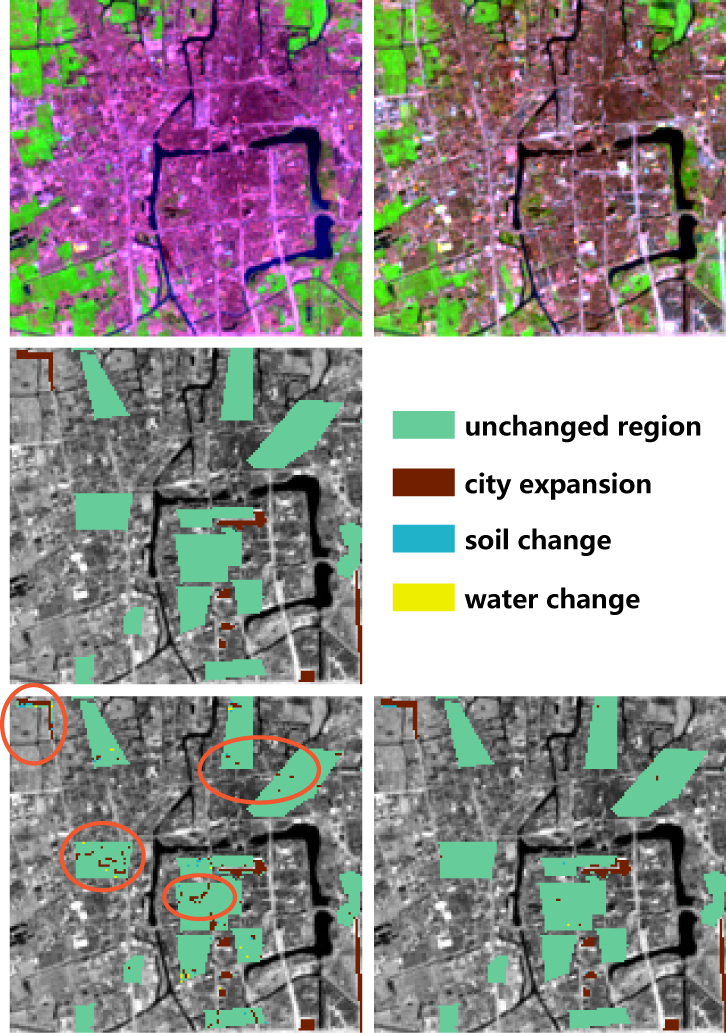}
\renewcommand{\figurename}{Fig}
\caption{\label{fig:label_noise} Comparison between spectral-temporal model (RNN) and spectral-spatial-temporal method (ReCNN-LSTM) on a region of the Taizhou city. From left to right and top to bottom: $T_1$ image, $T_2$ image, ground truth, change detection map obtained from RNN, and change detection map produced by ReCNN-LSTM. It can be clearly seen that there are a number of noisy scatter points of wrong detection (see ellipses in the lower left image) in the change detection map of RNN. While our spectral-spatial-temporal model RCNN-LSTM addresses this problem by eliminating those points.}
\end{figure}

\subsection{Analysis of Spatial Component: RNN vs ReCNN-LSTM}
In the case of spectral-spatial-temporal change detection, the proposed recurrent convolutional network is able to significantly improve the spectral-temporal-based RNN model. As shown in Table~\ref{tab:binary}, compared to RNN, ReCNN-LSTM increases the accuracy of binary change detection considerably by 2.23\% of OA and 0.0708 of Kappa coefficient, respectively, on the Taizhou data set. For the Eppalock lake scene, the accuracy increments on OA and Kappa coefficient are 3.46\% and 0.071, respectively. Table~\ref{tab:multi} compares the performance of RNN and ReCNN-LSTM in terms of multi-class change detection task. The latter can improve the former by 2.56\% of OA and 0.0905 of Kappa coefficient, respectively, on the Taizhou scene; by 2.36\% of OA and 0.036 of Kappa, respectively, on the Eppalock lake data. These results reveal the fact that the usage of spatial cue in our model can construct a more powerful spectral-spatial-temporal change detector.
\par
Furthermore, as shown in Fig.~\ref{fig:label_noise}, it is obvious that the spectral-temporal change detection method (RNN) always results in noisy scatter points in the change detection map. However, our spectral-spatial-temporal model ReCNN-LSTM addresses this problem by eliminating noisy scattered points of wrong detection.

\subsection{Comparison with Other Approaches}
The OAs and Kappa coefficients of all competitors and the proposed networks on binary change detection task can be found in Table~\ref{tab:binary}. The classical change detection algorithms, CVA, PCA, MAD, and IRMAD, all achieve a good performance, especially IRMAD, which has the best performance among them. Compared to IRMAD, improvements in OA and Kappa coefficient achieved by ReCNN-LSTM are 3.59\% and 0.1279, respectively, on the Taizhou data set, and increments of OA and Kappa obtained by ReCNN-LSTM on the Eppalock lake scene are 7.4\% and 0.1554, respectively. However, the cost of such accuracy improvements is that we have to manually label some training data for supervised learning.
\par
Table~\ref{tab:multi} presents accuracy indexes on multi-class change detection task. Analysis of the detection accuracies indicates that SVM with RBF kernel outperforms DT, mainly because the kernel SVM generally handles nonlinear inputs more efficiently than DT. It can be seen that the proposed recurrent convolutional network ReCNN-LSTM outperforms SVM and RNN in terms of OA and Kappa coefficient on both the Taizhou and Eppalock lake data. Compared to SVM and RNN, ReCNN-LSTM increases OA by 4.14\% and 2.56\%, respectively, on the Taizhou data set; by 2.84\% and 2.36\%, respectively, on the Eppalock lake data.
\par
Fig.~\ref{fig:change_maps} shows change detection results of the Taizhou city and Eppalock lake obtained by our model.

\section{Conclusion}
\label{sec:con}
In this paper, we have proposed a novel neural network architecture, called recurrent convolutional neural network (ReCNN), which integrates merits of both convolutional neural network (CNN) and recurrent neural network (RNN). ReCNN is capable of extracting joint spectral-spatial-temporal features from bi-temporal multispectral images and predicts change types. Moreover, it is end-to-end trainable. All these properties make ReCNN an excellent approach for multitemporal remote sensing data analysis.
\par
The experiments on real multispectral images demonstrate that ReCNN achieves competitive performance, compared with conventional change detection models as well as spectral-temporal-based RNN algorithm. This confirms advantages of the proposed recurrent convolutional network. In addition, ReCNN is a general framework; therefore, it can be applied to other domains and problems (such as multitemporal hyper/multi-spectral data classification) that involve sequence prediction in remote sensing sequence data.
\par
Future works will focus on new architectures based on ReCNN, for example, a semi-supervised ReCNN that can also use arbitrary amounts of unlabeled data for training -- typically a small amount of labeled data with a large amount of unlabeled data.

\section*{Acknowledgement}
The authors would like to express their appreciation to Dr. Chen Wu for providing the Taizhou data set.

\ifCLASSOPTIONcaptionsoff
  \newpage
\fi

\bibliographystyle{IEEEtran}
\bibliography{IEEEfull,ReCNN_v1}

\end{document}